# *The Algorithm of Islamic Jurisprudence (Fiqh) with Validation of an Entscheidungsproblem*

**Elnaserledinellah Mahmood Abdelwahab [a*], Karim Daghbouche [a], Nadra Ahmad Shannan [b]**

[a] makmad.org e.V., Hanover (Germany)

[b] Faculty of Shariah, Umm Al-Qura University, Makkah (Kingdom of Saudi Arabia)

Corresponding author: elnaser@makmad.org

**ABSTRACT**

The historic background of algorithmic processing with regard to etymology and methodology is translated into terms of mathematical logic and Computer Science. A formal logic structure is introduced by exemplary questions posed to *Fiqh*-chapters to define a logic query language. As a foundation, a generic algorithm for deciding *Fiqh*-rulings is designed to enable and further leverage *rule of law* (vs. *rule by law*) with full *transparency* and *complete* algorithmic coverage of Islamic law eventually providing *legal security, legal equality,* and *full legal accountability*. This is implemented by disentangling and reinstating classic *Fiqh*-methodology (*usul al-Fiqh*) with the expressive power of subsets of First Order Logic (FOL) sustainably substituting *ad hoc* reasoning with falsifiable rational argumentation. The results are discussed in formal terms of completeness, decidability and complexity of formal *Fiqh*-systems. An *Entscheidungsproblem* for formal *Fiqh*-Systems is formulated and validated.

**Keywords:** algorithm, *Khwarizmi*, Islam, balance, algebra, *Shariah*, Islamic law, jurisprudence, logic, syntax, semantics, *Entscheidungsproblem*, complexity, decidability, SAT-problem, transparency, security, accountability, rule of law, rule by law, *Fiqh, usul al-Fiqh*

## 1. INTRODUCTION

The here presented *Algorithm of Islamic Jurisprudence* connects back some 1,200 years to the origin of the modern term *Algorithm* which is the most fundamental methodological driver of the current information technology pace at internet speed.

It will be discussed in three major parts which consist of (A) methodological considerations in *Islamic Jurisprudence* (*Fiqh*) with regard to (B) Computer Science and its underlying mathematics, and a discussion (C) of the practical algorithmic results in computer programming while this introduction shall serve as a brief historic review:

It is commonly known that the term *Algorithm* is a Latin short-version of the name *Muhammad ibn Musa Al-Khwarizmi* (163-235AH / 780-850AD), who is the author of the Arabic book *Kitab Al-Jabr wa-l-Muqabala* (215AH /

830AD) [Al-Khwarizmi], i.e., *The Compendious Book on Calculation by Completion and Balancing*. This book was translated into Latin in the 12th century A.D. entitled *Liber Algebrae et Almucabola* [Chester] with *algebrae* and *Almucabola* being transliterated into Latin from the Arabic title where the term *Algebra* is derived from *Al-Jabr* in the title of *Al-Khwarizmi's* book.

While there is widespread belief that *Kitab Al-Jabr wa-l-Muqabala* is a textbook for mathematics which, among others, introduced general rules to solve algebraic problems with one variable reducible to quadratic equations, it is first and foremost a textbook for *Islamic Jurisprudence*:

As traditionally and practically done in *Islamic Jurisprudence* (*Fiqh*), the first half of *Kitab Al-Jabr wa-l-Muqabala* introduced the applied methodology and term definitions, i.e., then as now the domain of *Algebra*.

The remaining half of his book solves legal questions on trade (commercial transactions), geometry (plane surface distributions) as well as testimonies.

Based on the newly introduced algebraic method, the by far most important part of *Kitab Al-Jabr wa-l-Muqabala* deals with Islamic heritage law which is complex with regard to as well as both, number of variables and number of *Fiqh*-rules (axioms).

Accordingly, *Al-Khwarizmi's* Algebra is just what its author says in the introduction: "[a] work on algebra, confining it to the fine and important parts of its calculations, such as people constantly require in cases of inheritance, legacies, partition, law-suits, and trade, and in all their dealings with one another, or where surveying, the digging of canals, geometrical computation, and other objects of various sorts and kinds are concerned." [Al-Khwarizmi]

It is evident that the introduction of a new *Fiqh*-method back then as well as today requires a standard set of persuasive arguments in terms of reproducible propositions and proofs.

Therefore, every *Fiqh*-problem in *Kitab Al-Jabr wa-l-Muqabala* is reducible to a linear equation while the more fundamental method of quadratic polynomials is never used but proven.

But before discussing the theory of quadratic polynomials (cf. [Al-Khwarizmi]), it was necessary to conclude some preliminary research such as introducing the calculation with Hindu numerals including the number zero with *Kitab Al-Jam wa-l-tafriq bi-hisab Al-Hind* (The Book of Bringing together and Separating According to the Hindu Calculation - lat.: *Algoritmi de numero Indorum*) written about (210AH / 825AD), i.e.: *Al-Khwarizmi* developed and published the Hindu Calculation half a decade before introducing the new *Fiqh*-method of reducing variables to quadratic equations.

While *Al-Khwarizmi's* proposition consisted of reducing **selected** *Fiqh*-problems to linear equations, the here introduced *Algorithm of Islamic Jurisprudence* proposes to cover the **complete** *corpus juris* of *Islamic Jurisprudence*.

However, the most important proposition underlying the *Algorithm of Islamic Jurisprudence* consists of one single statement:

    There exits a complete legislation.

Its mathematical and algorithmic formalization will be presented in part B of this paper.

This statement implies that every *Fiqh*-problem has a solution, i.e., that every religious case has a ruling,[1] which is the traditional conviction from its inception:

*Fiqh* developed in two stages: Firstly during the time of Prophet Muhammad's (saw)[2] companions when *Fiqh* did not rely on explicit rules. Rather, it relied on *synthetic* understanding by the Prophet's (saw) companions as they witnessed the revelations of the Quran with a situation- and procedural awareness of Islamic law.

As the Islamic influence and responsibility grew, the Muslim community was embraced by diversity as well as both, culturally and intellectually. At the same time, the personal connection of *Fiqh* to the Prophet (saw) and his companions became more and more remote. This let *Fiqh* to be studied *analytically*. It also required in-depth studies of the Arabic language and logic that were relevant to the study of the major sources of Islamic law, namely the Quran, the Sunnah, *ijma'*, i.e., the unanimity of the community, and *ijtihad*, i.e., the scholarly discretion.

*Ijtihad*, which means the exercise of scholarly endeavor to generate the right rulings in different situations, started during the lifetime of Prophet Muhammad (saw), but it was limited in scope and remained within the area of personal matters, mainly with regard to transactions and commitments (cf. [Al-Dawalibi]). After the Prophet (saw) had passed away, his companions produced rulings on different matters, *synthetically* relying on their personal discretion. Their rulings were of three types:

A) interpretation of the Quran and the Sunnah[3]
B) analogy that draws on similarities between cases (i.e., *qiyas*), and
C) discretion that does not rely on any particular text

When the major works on *Fiqh*-methodology, such as *Al-Shafi'i's Al-Risalah* (204AH / 820AD) [*Sahfii*] and Al-*Ghazali's Al-Mustasfa* (504AH / 1110AD) (cf. [Hammad]) appeared, scholars had different views on the validity of the second type of scholarly discretion, i.e., on *analogy*.

Many of the works on *Islamic Jurisprudence* that discuss analogy include questions concerning logic[4]. Other scholars who deny the relevance of logic to *Islamic Jurisprudence* consistently rejected the very concept of analogy. They deprecated the introduction of methods that are alien to the genuine, Arab Muslim mind.

Nevertheless, a number of leading scholars of *usul al-Fiqh*, such as *Al-Ghazali* and *Al-Razi* (240-312AH / 854-925AD) opposed this position (cf.

---

[1] Its inversion (complete induction) would simply imply that every religious ruling has a reason.

[2] saw = sall Allahu 'alay-hi wa-salaam = may God pray on him and grant him peace (a traditional saying after mentioning his name)

[3] *Sunnah* is a general term that refers to the Prophet's (saw) example and guidance, whether verbal or practical.

[4] Their understanding of logic was solely based on the well-known syllogisms of the Greek which were studied in the – then new – context of *Fiqh* resulting in selections of appropriate "ways of analogy". Muslim scholars invented the idea of expressing syllogistic methods and their explanations is poetic form as can be found in, e.g.: Al-Akhdary – *Al sullam al Munawraq fil Fiqh* (cf. *Al-Akhdari*. (2014, March 26). In Wikipedia, The Free Encyclopedia. Retrieved 21:36, April 7, 2014, from http://en.wikipedia.org/w/index.php?title=Al-Akhdari&oldid=601304682)

[Hammad]), citing evidence from the Quran and the Sunnah to confirm the validity of analogy as a source of rulings on questions that are not directly addressed in the Quran or the Sunnah.[5]

## A. FIQH-METHODOLOGY (*USUL*)

Our purpose in this part is to present some questions of *Fiqh*-methodology (hereafter referred to as Arabic: *usul al-Fiqh*)[6] and explain their relevance and relation to logic and algebraic methods in general. This will serve as preliminary to the contemporary trends in logic adopted by most experimental and mathematical sciences to mechanically (algorithmically) determine whether conclusions are true or false. The logical trend had its roots in the Muslim scholarly mind that realized that in dealing with religion, it could not exclusively rely on pure logical reasoning to draw true conclusions. It required two other essential elements, namely language and law. This implies that *logic is a subset* of *Islamic Jurisprudence*.

We need to make certain definitions in order to distinguish between the three disciplines: language, jurisprudence and logic:

1. A Logical Question (Qgic) is a *syntactic* problem that cannot be solved except through a mental link between the realities of the outside world and the natural human concepts, which is totally independent of agreed terminology and belief. (cf. [Hammad])
2. A Language Question (QLn) is a *semantic* problem that is solved by reference to the meanings of the linguistic patterns it uses, whether these are words or expressions and sentences. These meanings are either essentially or traditionally known in Arabic.
3. A *Source* Question (SF) is a search and authentication problem, i.e., an *algorithmic* problem that can only be solved by reference to a finite set of rulings that have been approved on the basis of Quran and Sunnah.

With these terms at hand we may further address and classify methodological questions such as analogy, which is the most important of all:[7]

Analogy, or *qiyas*, denotes the analytical reasoning process that scholars of *Islamic Jurisprudence* use to work out *Fiqh*-rulings. It relies on a finite set of logical rules for deductive reasoning.

As such, *qiyas* represents the main junction of *Islamic Jurisprudence*. Because although the deductive process of *qiyas* is truth-preserving if applied correctly on true axioms, there may always be debate concerning the main and secondary postulations used prior actual deduction and whether these postulations were deduced correctly from religious texts.

---

[5] cf. *Al-Ghazali's* chapter on answering the arguments denying the validity of analogy in: [Hammad]

[6] *Fiqh* is an Arabic term meaning "deep understanding" or "full comprehension". Technically, it refers to the body of Islamic law extracted from detailed Islamic sources (which are studied in the principles of Islamic Jurisprudence) and the process of gaining knowledge of Islam through jurisprudence. The historian *Ibn Khaldun* (732-808AH / 1332-1406 AD) describes *Fiqh* as "knowledge of the rules of God which concern the actions of persons who own themselves bound to obey the law respecting what is required (*wajib*), sinful (*haraam*), recommended (*mandub*), disapproved (*makruh*) or neutral (*mubah*)" [Glasse]. The Arabic term *usul* means lit. *root* denoting *methodology*. This definition is consistent amongst Islamic jurists.

[7] There may be more detailed definitions while the here presented is sufficient for clarification.

The issue that causes such debate is *semantics*, i.e., the intention and purpose of the speaker materialized as a dedicated linguistic expression, i.e., *syntax*.

This debate is continuous because any investigation about semantics is generally characterized by not restricting the analysis to linguistic expressions as such and their relations to one another. Rather, those analysis take into consideration what is being referred to in the sentences of the object language. And that is independent of the method chosen (i.e., the method of denotation relation (cf. [Frege]) or of extension and intension (cf. [Carnap]) because there is always the range of those objects taken into consideration to which the expressions of the object language relate. All specific semantic predicates such as "denotes", "term xyz-denotes", "satisfies", "true", etc. provide directly or indirectly (the latter as in the case of the predicate "true") with relations between linguistic expressions and their semantics.

Going one step further by not only abstracting away the semantics of the speaker who uses the expressions of the object language, but by also abstracting from what the linguistic expressions refer to, yields a purely formal analysis where any semantics are replaced by the *syntax*. Hence, in the most abstract sense we return to the fact that logic can operate purely syntactically, that is:

The basic concept of logical deduction is exclusively defined syntactically (as opposed to the concept of "truth"). While a formal consideration of an expression doesn't provide knowledge about the speaker's purpose or whether the sentence is true or not because this knowledge goes beyond that to which a sentence refers to, this knowledge is not required for the logic deduction *per se*. In a precisely structured language system, it can be decided whether two submitted expression can be derived immediately, i.e., logically deduced one from the other or not (or any of several other submitted expressions) without reference to their meaning provided that the term "immediately derivable" has been clearly determined in the system in question.

Therefore, any derivation, no matter how long or complex, can be characterized syntactically (cf. [Stegmüller]), i.e., the deductive part of Shariah (*qiyas/analogy*) could be mechanized completely, which takes us back to the origins of the term *Algorithm* with a logical connotation. (cf. [van der Waerden])

*Usul al-Fiqh* discussed this finding extensively, provided with rules and classifications according to their validity, relevance, correct application to the view in question as well as how balanced they interpret the relevant text. At the same time, the Arabic vocabulary was investigated and classified into a couple of main sections, the most important of which is the one concerned with *the general* and *the specific*. A detailed study of semantics was carried out classifying word meanings into ones that are specifically and expressly intended, and ones that are inferred from the text itself. The reasons were analyzed with regard to different attributes, including simplicity, complexity, limitation, extension, prevention and nullification. All these questions are purely logical (Qgic). The essential purpose of such studies has been the establishment of accurate criteria to ascertain the most balanced religious verdict.

## 2. THE GENERAL AND THE SPECIFIC

The emphasis placed on the relation between *syntax* and *semantics* is reflected in studies on the general. It discusses in great detail the reasons for using the general linguistic formulae from the point of view of the individual meaning that the general term includes. It seeks to answer questions like: Does the text mean the least group? Does the general term have a meaning that applies universally or specifically? Does the general term apply to all individuals? Obviously, all these questions are common to language, whether natural or formalized such as algebraic logic (QLn & Qgic).

*Fiqh* also provides with studies about the case of the individual that is included in a general term. It was realized that the individual may be referred to without using any general linguistic formula. In such a case, the reference is made through the usage of a description that applies to the individual without further details. The reference may at times be restricted by further descriptions the text itself does not include (QLn & Qgic).

It was also concluded that reasons for a verdict are, in the majority of cases, closely linked to how the universal is related to the restricted case of its kind (Qgic and SF).

One of the prime *Fiqh*-examples is the punishment specified for two offences, namely, a vow by a husband making his wife unlawful to him (*zihar*) as well as manslaughter.

The punishments for both are mentioned in the Quran and include the freeing of a slave, but the relevant verses mentioning the crime of manslaughter specify that the slave to be freed should be a Muslim male or female slave, while the verse mentioning the punishment for *zihar* speaks about freeing a slave without any further specification. *Fiqh*-scholars conjugated the two cases and concluded that the male or female slave to be freed in both instances must be a Muslim because both actions are forbidden. This reason, i.e., the prohibition of both actions, represents the highest degree of abstraction of the causes (Qgic).

## 3. SEMANTICS AND REASONS

Studies of words were not limited to their general linguistic or logical meanings. These studies were expanded to include the specifically intended meanings and those that are implied and understood through a pointer given in the text (Qgic), i.e., through a syntactic pre-differentiation.[8]

These studies provide with details about specifically intended meanings so as to include what is required to ensure that a given statement is correct, logically and religiously, while the meanings understood through a pointer were divided into what is understood through being consistent or being inconsistent (Qgic).

The studies of reasons unveil a unique and detailed interaction between logic

---

[8] It has to be taken into account that there is no syntactic decision procedure which would allow our thinking to semantically relate morphologically different types of terms to each other because the understanding of the meaning of a sentence in natural language is the same as knowing its truth conditions. (cf. [Eley])

Since we only have an explicit grammar (syntax) in terms of the ontological-declarative *Shariah*, we need to perform a pre-differentiation of cognitively tangible facts (perceivable facts), which indeed renders the axiomatic, i.e., the explicit character of the facts, unattainable for *Shariah*. The mere hypothetical nature of *trial and error*, however, is conditioned by a heuristic maximal-approximate predicate (cf. Figure 2 and [Daghbouche 2012]) and language, considering, that these are usually mere questions of logic.

For illustration, we will discuss the reasons of prevention and nullification, as well as the reasons derived from attributes, action, and denotement.

Scholars of *usul* consider the reason to be a clear and accurate description that may be real (Qgic), based on tradition (QLn), linguistically (QLn) or religiously (SF), while it could also be non-existent (Qgic). (cf. [Hito])

Thus, the reasons of prevention and nullification are clear and balanced descriptions that occur during the performance of an action and nullify its verdict (nullification), or before it is undertaken to prohibit it (prevention). As such, they are related to a negative ruling.

An example is prayer and invalidation of ablution (*wudu*), such as urination: It prevents starting prayer if it occurs before it, stops prayer and nullifies it if it occurs during it. Another example relates to marriage when the two partners follow different religions. If one of the parties follows a religion other than the three monotheistic book religions, marriage to a Muslim is not permitted. If this happens during an existing marriage, it nullifies it (after the end of a waiting period). The point here is that although urination and the difference of religion are situations that have clear religious descriptions (SF), giving them the status of preventive and nullifying reasons is a verdict of logic. It is based on either inductive or deductive reasoning according to the different scholarly approaches to rules and differentiations.

Reasons derived from terms become operative when the language is ascertained through analogy. Thus, whatever *Arabs* denote with *and* attribute to "wine" is forbidden to drink. This is one of the most fundamental linguistic questions and many scholars are not in favor of using analogy through language (QLn).

## 4. ANALOGY

Analogy or *qiyas* is a deductive process. Analyzing different aspects of analogy, the most important ones include *certainty*, i.e., the need for generating a verdict on a secondary question and the clarity of the reason on which a verdict is based.

All these are linguistic and religious questions (QLn & SF). An absolute analogy cannot be obtained without maximum certainty of the reason for the ruling on the primary question and maximum certainty that this very reason applies to the secondary question[9].

For example: It is explicitly forbidden to even say the word "Ugh" to one's parents. Scholars have ruled through analogy that it is also forbidden to beat one's parents. It is certain that the reason for prohibiting to saying "Ugh" to one's parents is that it hurts them. Accordingly, it is assumed that beating them causes greater hurt. However, we cannot maximize certainty in all this without awareness about the meanings of the terms: *beating, ugh*, and *hurt*.

In terms of logic we note that analogy is divided from a purely logical point of view into *inverse* and *direct* analogy. This is based on the premise that the verdict of the primary case is either positive or negative in relation to the verdict on the secondary case.

If it is positive, i.e., the same verdict as the verdict on the secondary case, it is called *direct*; otherwise it is *inverse*.

---

[9] The idea of having a solved primary questions which forms the basis for the analytical decision regarding a secondary one can be traced back to the time of Greek logic.

Thus the rule of inverse analogy is:

If the reason of the primary case leads to its verdict, the presence in the secondary case of the opposite of this reason leads to the opposite verdict on this case. This may formally be represented as:

<u>Fs primary ↔ R primary</u>
↔ *Inverse* (Fs secondary)
↔ *Inverse* (R secondary)

The analogy is not valid unless the reason is a necessary *and* sufficient condition for giving a verdict on both, the primary and secondary case.

We will exemplarily apply this rule to a case cited in books of *usul*:[10]

> If a person pledges to perform the practice of I'tikaf[11] and fast, the fasting is a necessary condition for the validity of one's I'tikaf according to all scholars, but if one pledges to perform the I'tikaf and pray, combining both acts is not necessary for the validity of one's I'tikaf. Scholars differ as to the condition of fasting during the I'tikaf if the concerned person has not made a pledge. *Abu Hanifa* requires fasting in all cases, whether pledged or not. *Al-Shafi'i* does not require fasting as a condition. *Abu Hanifa* gives the following formula of inverse analogy: On the basis of analogy with prayer: had fasting been unnecessary for the validity of I'tikaf when it is not specified, it would not be a necessary condition when it is specifically pledged. In the case of prayer, as it is not a necessary condition when not specified, it remains unnecessary when pledged. Thus, taking prayer as the primary case, the established verdict is that it is not a necessary condition for the validity of I'tikaf with the reason being that it is not obligatory when pledged. The established verdict on the secondary case is that fasting is a necessary condition for the validity of I'tikaf with the reason being that it becomes obligatory when pledged. The two cases differ in ruling and reason.

To represent the question formally we will make the following abbreviations:

Fs ≡ I'tikaf done
Prv ≡ prayed consistently during I'tikaf
Pgv ≡ pledge performed before I'tikaf
Fv ≡ fasting observed during I'tikaf

A → B ≡ B necessary condition for A
*Inverse* (A) ≡ the predicate (A) inversed
A ∧ B ≡ the predicate A added to B
A ∨ B ≡ the predicate A or B is correct

As given in the book, the case may be presented as follows:

(Fs ∧ Pgv → Fv) → (Fs → Fv) in the secondary case
And the *inverse* ((Fs ∧ Pgv → Prv)) → *inverse* (Fs → Prv) in the primary question.
The analogy goes: It is well known that (Fs ∧ Pgv → Fv), so (Fs → Fv) is arrived at through the inverse of the reason of the primary case and by substituting Prv by Fv.

However, the presentation of the secondary case is incomplete. It is not possible to logically deduce (Fs → Fv)

---

[10] cf. [Al-Ghazali: p. 140] and [Hito: p. 375]

[11] The I'tikaf is a recommended practice during the final days of Ramadan, when a person performing it spends days in a mosque, going out only for the necessary reasons, and devotes all his wakeful time to worship in various forms.

from (Fs ∧ Pgv → Fv) because the first rule is more general. The fact is that the reason in the primary case is not only:

*Inverse* (Fs ∧ Pgy → Prv)

Had it been so, it would not have been possible to use this analogy because this reason is not necessary and sufficient for:

*Inverse* (Fs → Prv)

The correct reason is:

*Inverse* (Fs ∧ Pgy → Prv) ∧ *inverse* (Fs ∧ inverse (Pgv) → Prv)

This means that prayer is not essential for the validity of the I'tikaf whether pledged or not. Hence the secondary case is:

[(Fs ∧ Pgv → Fv) v (Fs ∧ *inverse* (Pgv) → Fv)] → (Fs → Fv)

In plain language:

If fasting is an essential condition for the validity of I'tikaf, whether pledged or not, then it is an absolute condition. However, we know that it has been stipulated in a pledge and as such it is an absolute condition.

This example implies two heuristic principles:

Firstly, a commitment to logical analogy provides with a well-considered and reflected verdict. Although the discussed *Fiqh*-question was discussed in an incomplete or not perfectly correct way in ancient books, *Abu Hanifa's* view is correct according to the rules of logic. Therefore, *Shafi'i* scholars need to find a valid argument to refute the *Hanafi* verdict.

The second heuristic principle is the adoption of logic in formalizing the *Fiqh*-question with the process of deduction making the mental process and arguments leading to a verdict *transparent*, i.e., fully reproducible.

## 5. DEFINITIONS OF RULES FOR USUL AL-FIQH AND FIQH

At the beginning of the second stage of *Fiqh* a debate started between scholars of the *Hijaz*[12] who adhered to the literal interpretation of the texts and the scholars of *Iraq*, who had limited knowledge of Hadith but were highly skilful in analytical reasoning. The debate motivated leading scholars to establish rules of *usul*. This complex task was undertaken, among others, by *Al-Shafi'i* with his book *Al-Risalah*.
Eventually *Fiqh* was performed by two separate methods with the *Hanafi*-school frequently relying on analogical deduction and independent reasoning, and *Maliki* and *Hanbali* generally using the Hadith instead where *Shafi'i*-school uses Sunnah more than *Hanafi* and analogy more than the two others.
The first method was characterized by leaning towards analytical reasoning as much as possible, discussing primary questions in isolation of secondary ones. Secondary questions in turn must be subject to treatment of the rules of *usul* but could not act as logical premises unless such attribution was justified by separate evidence, e.g.:
Quranic imperatives such as "Attend prayers regularly" imply questions on whether the imperative form specifies a duty applicable to all adults, or if it could be understood in terms of a recommendation. On the basis of grammatical rules and basic principles of the Islamic faith, i.e., common sense (QLn and SF), *usul*-rules established that a statement in the imperative form specifies an obligation, generating two aspects of evidence: a general one, i.e., an obligatory order, and a specific one, i.e., an order to offer prayers regularly.

---

[12] Literally "the barrier" primarily defined by the cities *Makkah*, *Medina*, and *Taif* in the current "Kingdom of Saudi Arabia"

The combination of these two generate a *Fiqh*-ruling (verdict): "Prayer is obligatory".[13]

The second method establishes the *usul*-rules on the basis of the rulings given by leading scholars of the *Hanafi*-school of thought for secondary cases. Scholars adopting this method undertook an extensive study of the rulings given by their leading predecessors, induced rules and applied them to various questions. If they found their conclusions conflicting with the rulings given by their predecessors, they amended the rules so as to provide consistency between the two, or else they excluded specific questions from their rules. Depending on the field of application, there are two major categories for rules which concern *usul* and *Fiqh*. The following will provide with some preliminary definitions which will be further differentiated in the course of the then following chapters:

*A rule of usul* is a general verdict that is used to identify the elements of *Fiqh*-evidence,[14] how it may be used,[15] and the status of the person to whom to apply.[16]

*A rule of Fiqh* is a practical component of *Islamic Jurisprudence* that is used to identify the rulings applicable to its details. If these details come under different themes of *Fiqh*, the rule is a general one, and if they come under one section, it is a specific rule, e.g.:

"If the verdicts of permissible and forbidden apply in one case, the forbidden takes precedence." Under this rule, the following implications apply:

1. If one part of a tree is in the *Haram* area and another part is in the *Hil*[17] area, the tree is forbidden to fell.
2. If a Muslim butcher and an idolater butcher take part in slaughtering an animal, that animal is forbidden for consummation.
3. If meat of an animal slaughtered according to the Islamic tradition is mixed with carrion, nothing of it is permissible to eat.

---

[13] Another example is the rule that says: "A noun with an indefinite article implies general application when it occurs in a context of negation." This rule is generated from the Quranic verse that says: "This is the Book, there is no doubt about it." [Quran 2:2] When this rule is applied, the verse means that no type of doubt can apply to the Quran.

[14] Such as "unanimity is a valid source of rulings"; "An order signifies obligation while a negative order signifies prohibition".

[15] In other words: how rulings are to be derived from it when there are conflicting or equally valid considerations, such as: "The text takes precedence over the apparent state; a report transmitted by many reporters at every stage is given precedence over one reported by single transmitters; the unspecified is explained in the light of the restricted, and the general in the light of the specific, etc."

[16] This applies to a scholar who is competent to exercise scholarly discretion, i.e., *ijtihad*, and one who follows a particular school of thought, and the criteria applicable to each.

[17] The *Haram* area is the area surrounding *Makkah* extending a few km in some places and up to 20km in others. It is subject to certain restrictions, including the prohibition of felling its trees. The *Hil* area is the complement of the *Haram* area.

## 6. FIQH AND LOGIC

Having briefly discussed *usul*-rules and how they were anticipated, a closer look at the rules of *Fiqh* may shed light on their practical signification and their relation to logic.

*Fiqh*-rules were developed organically. Once reaching maturity, scholars studied the foundations with its rules drawing analytic comparisons between different schools of thought,[18] including:

1. Looking into *Fiqh*-questions in a regulated way and to facilitate their memorization. The rules were formulated in easy and precise language minimizing room for confusion and making it easy to memorize. Examples are: "What is permitted for a specific cause is no longer permitted when the cause is inapplicable"; "There shall be no infliction of harm on oneself or others"; and "Necessities permit what is forbidden".
2. Development of *Fiqh*-insights by enabling the grouping of questions with similar aspects as well as separate dissimilar ones.
3. Understanding the objectives of *Islamic Jurisprudence*. Knowledge of a general rule that applies to a large number of questions giving a clear view of the objectives of the legislator. This, e.g., motivated *Al-Izz ibn Abd Al-Salam* to include the whole *Islamic Jurisprudence* under the major rule of "Bring benefit and prevent harm". *Al-Subki*, e.g., even abstracted further suggesting the rule simply to be "Bring benefit", because the prevention of harm is already implied in this formula.
4. Rules as basis to look at questions having similar reasons which were not given rulings or discussed by earlier scholars.

The last point is the one that is particularly important for our present discussion, as it means that contemporary cases can be examined on the basis of these rules. Undoubtedly they cannot be put together unless there are plenty of secondary questions that have similar bases of deduction. When researching the references outlined in old books on subject matter we observe:

Firstly, most books list *Fiqh*-rules together with *usul*-rules and *Fiqh*-criteria. The reason may be that it was not possible to practically separate these two sets of rules. Moreover, the definition of *usul*-rules is too broad and lacks precision. Besides, they apply to many *Fiqh*-rules as well. To give an example according to this definition, the rule that requires reference to both, habit and social tradition, would be an *usul*- and *Fiqh*-rule at the same time.

Secondly, most scholars who list such rules do not cite evidence in support of them because most of the rules were set by quasi-logical induction.[19]

The inductive process uses detailed or particular cases to infer a general rule that may be complete or incomplete.

Hence, we suggest drawing the line of demarcation between *usul*-rules and *Fiqh*-rules in a mere formal sense, i.e.:

*Usul*-rules are finite, denumerable, and constant sets of axioms inductively deduced from source texts (QLn). They treat (in almost all cases) *semantics* such as the meaning of a statement using the imperative form, or the meaning of a general/specific statement. As discussed earlier, we denote linguistic questions with QLn attributing to be generic expressions and serving as deductive bases, i.e., as axioms.

---

[18] This is a natural development that applies to most modern natural sciences.

[19] i.e., it pretends logic reasoning while it is mere *ad hoc* decree.

*Fiqh*-rules look at the causal reasoning behind verdicts with a variety of *Fiqh*-details and exemptions provided with logic variables (Qgic). In anticipation of the implications for mathematical logic and as a summary of the foregoing chapters, it is important to note that logic variables of Qgic are pre-differentiated by QLn. This implies not only Qgic to be finite and denumerable, but also the possible set of values.

In this formal relation *usul*-rules act as axioms for detailed rulings with the *Fiqh*-rules processing the logic implications to accomplish the axioms (see Figure 1 and 2). Thus, if analogically the axiom "order signifies obligation" is set, every case that includes an order implies an obligation, e.g.:

Methodologically, one may find Sunnah texts such as: "Pray as you have seen me pray", "Recite the Quran", "Order your children to start praying when they are seven years of age", "If a fly falls into your plate, immerse it all", or "If a dog licks your plate, wash it seven times, one of which should be with dust and water." These texts are imperative in terms of *pray*, *recite*, *order*, *immerse* and *wash*. If the axiom *order signifies an obligation* applies all implications in terms of imperatives are mandatory actions (vs. voluntarily actions). If there are similarities between the reasons calling for such imperatives, as with the cases of immersing the fly and washing the plate, specific logic formulas can be formalized (variables),[20] e.g.:

"Paying zakat in advance of its due date", "Paying zakat-al-fitr (which is due at the end of Ramadan) at anytime during Ramadan", "Paying the compensation of accidental killing once the victim is injured, but before his death", and "Paying the atonement for the violation of an oath after saying the oath but before the violation."

Since these financial matters are all due if one of two different reasons apply,[21] a generalized rule would look like:

"A financial obligation which can be triggered by two different reasons may be settled even if only one of the two reasons applies."

In the first example, the relationship between the axioms "order signifies obligation" and "there shall be no infliction of harm on oneself or others" constitutes a hierarchy i.e., in the cases of a dog licking a plate and a fly falling into it, the second axiom provides an implicit reason for the first one.[22]

That finding yields two *heuristic propositions* formulated with regard to mathematical logic terms which will be deepened in the next section:

**Proposition 1:** In *Islamic Jurisprudence*, there exists a *Fiqh*-rule for the deduction of any legal verdict, deduction being done using a formal logic system, i.e., *Islamic Jurisprudence* constitutes a *Consistent Legislation*.

The corresponding concept in mathematical logic is *correctness* which is the ability of a formal system to deduce new assertions which are *consistent* within given axioms;

---

[20] The axiom common to both, the immersing and washing is: "There shall be no infliction of harm on oneself or others", which makes it a logical *Fiqh*-rule.

[21] The two reasons making the payment of *zakat* due are having more than the threshold and a year is completed.

[22] The other cases may be attributed to the axiom "bring benefit" because prayer and the recitation of the Quran provide benefits.

**Proposition 2:** For every query/legal *Fiqh*-question there exists a legal verdict, i.e., *Islamic Jurisprudence* constitutes a *Complete Legislation.*

The corresponding concept in mathematical logic is *completeness* which is the ability of a formal system to produce a proof of any assertions related to the domain of discourse.

**Lemma 1:** If proposition 1 and 2 both proof valid, *Islamic Jurisprudence* would be *complete* with respect to a set of given *Fiqh*-axioms and an intrinsic formal system of deduction.

**B. LOGIC & COMPUTER SCIENCE**

Since the inception of algorithmic processing, namely with computer science in the early 40ies of the past century, it was realized that any technology to automate logical inferences could have tremendous potential to solve problems by drawing automated conclusions from any given fact. Up-until-today, the most powerful and expressive formal method for describing and analyzing information is represented by First Order Logic (FOL). In this abstract and formal context, logic shall just be the concept of what follows from what, e.g., if two statements x, y are true, then one can infer a third statement z from it while it doesn't matter if any of the statements is actually true! The formal, i.e., *syntactic* quality is simply that if x and y are true, then z must also be a logically valid statement.

## 7. MATHEMATICAL LOGIC AND FORMAL LOGIC

As a brief overview, far from being formal, mathematical logic comprises of two constitutive features for the here presented:

Firstly, it intends a complete and concise formulation of formal logic (incl. FOL)[23] in so far that complex mathematical theorems can be expressed in simple, singular formal propositions. (cf. [Whitehead/Russell])

This has the reductionist advantage that secondly, mathematical axioms in arithmetic, algebra, geometry, etc., can be described by countable many logical calculi in terms of statements about classes, relations, and syntax. (cf. [Frege])

In this sense, the second constitutive feature of mathematical logic according to the "western"[24] understanding was first understood by *Leibniz* (1646-1716) in his "*Characteristica Universalis*". (cf. [Gerhardt])

The symbolism, the hierarchy of classes of statements, the syntactic linking of rules, simply, the properties of a system of concepts constitutes all necessary areas of human mental activity, which in turn, using this very system, becomes transparent. (cf. [Wittgenstein] and [Tarski])

Thus, the constitutive role of mathematical logic may be introduced as preceding all other sciences, or in the words of *Leibniz*:

"…that humanity would have a new kind of an instrument increasing the

---

[23] More specifically, the axiomatic set theory. The classes of formalized areas are: propositional logic, predicate logic with quantifier (incl. syllogistic), predicate logic with identity and identity theory [cf. Daghbouche 2013: p. 68]

[24] "western" is hereinafter referred to as connotation for "non-Islamic"

powers of reason far more than any optical instrument has ever increased the power of vision." [Whitehead/Russel: pp. XXVIII]

For the furtherance of our discussion it is largely irrelevant whether mathematics is regarded as a further development of logic (Logicism, represented by *G. Frege, B. Russell, R. Carnap*) or if it consists of calculi that are formed out of formal systems by preceding axioms using inference rules corresponding to theorems (Formalism, represented by *D. Hilbert, W.v.O. Quine, H.B. Curry*) or whether mathematics represents basic mental processes where the critical path consists of what can be constructed effectively due to these processes, but not what was raised as object of observation by the mathematician (Intuitionism represented by *L. Brouwer, A. Heyting, L. Wittgenstein, Lorenzen*).

In order to provide with sound foundations for mathematics, logicians of the 19$^{th}$ Century AD pursued the construction of a special formal system:

## 8. THE STRUCTURE OF LOGIC SENTENCES

This formal system, as a whole, represents an artificial language that is similar to natural languages in having a system of spelling and grammar that identifies possible sentences, distinguishing them from wrong expressions. For example, we cannot say in English, "A tree and ran I", because the conjunction "and" cannot occur between a noun and a verb. Likewise, the expression (FS → → F*v*) is meaningless because it contains an error of symbols, as the arrow symbol (denoting material implication) cannot be used twice in succession.

The structure of formal logic languages, particularly the simple ones that are intended for the study of mathematics, should include five sets of symbols[25]:

1. Symbols of logical statements (QLn);
2. Symbols of conjunctions (Qc);
3. Symbols of variables (Qv);
4. Symbols of rules (Qg);
5. A set of grammatical and dictation rules that distinguish structurally sound sentences for erroneous ones (Gmr).

Furthermore, we may assign the symbol (Sm) to indicate an infinite set of *Cartesian products* of the components of the sets QLn, Qc, and Qv. This yields a set (power set) which includes all possible relations between sets and their elements, independent of their magnitude.

Thus Sm represents the set of all possible sentences, correct or incorrect, that result from the use of the symbols of QLn.

Obviously, Sm is infinite even though the sets used to generate it are finite. This observation stresses the regulatory importance of Gmr because without proper procedures of how to combine sentences, the formal system would even be structurally infinite.

We will illustrate this with the formalized *Fiqh*-example (see section 4), which contains:

QLn ≡ (Fs, Prv, Fv, Pgv) and Qc ≡ (→, ∧, ∨, *inverse*)

If we abstract primary and secondary cases we realize that the following two rules apply:

---

[25] The description of the structure of languages used here is semi-formal and intends only to convey the main ideas of the topics on hand without dragging the reader into the burden of precise formalizations as those formalizations can be found in any textbook of mathematical logic.

$((Fs \wedge Pgv \rightarrow X) \vee (Fs \wedge \textit{inverse}\ (Pgv) \rightarrow X)) \rightarrow (Fs \rightarrow X)$

and $(\textit{inverse}\ (Fs \wedge Pgv \rightarrow X) \wedge \textit{inverse}\ (Fs \wedge \textit{inverse}\ (Pgv) \rightarrow X) \rightarrow \textit{inverse}\ (Fs \rightarrow X)$

X represents a variable that admits both, Prv and Fv. Therefore $Qv \equiv X$

The set of grammatical rules G*mr*, which is not explicitly mentioned in our example because we assumed its presence implicitly, may include for example QLn x Qc x QLn[26].
This means that an item of QLn is processed together with an item of Qc and another item of QLn following it; another example of Gmr might be: (QLn x Qc x QLn) x (Qc) x (QLn x Qc x QLn)

These formulae mean – for example - that we allow only expressions such as:

$(Fs \rightarrow Fv)$ or $(Fs \rightarrow Pgv) \rightarrow (Fs \rightarrow Fv)$

The deduction in the example of section 4 can thus be formally represented as:

Sen. 1: $(\textit{inverse}\ (Fs \wedge Pgv \rightarrow X) \wedge \textit{inverse}\ (Fs \wedge \textit{inverse}\ (Pgv) \rightarrow X) \leftrightarrow \textit{inverse}\ (Fs \rightarrow X)$
$\Leftrightarrow$ Sen. 2: $((Fs \wedge Pgv \rightarrow X) \vee (Fs \wedge \textit{inverse}\ (Pgv) \rightarrow X)) \leftrightarrow (Fs \rightarrow X)$

We replace X with Fv so that it becomes: $((Fs \wedge Pgv \rightarrow Fv) \vee (Fs \wedge \textit{inverse}\ (Pgv) \rightarrow Fv)) \leftrightarrow (Fs \rightarrow Fv)$

As we know that Sen. 3: $(Fs \wedge Pgv \rightarrow Fv)$ is realized.

Hence $(Fs \rightarrow Fv)$     (Q.E.D.)

Usually this set contains two types of rules: (a) Axioms of the formal system used (Qg') and (b) Axioms of the logical theory processed (Qg''). The classical syllogistic system traditionally used by ancient Muslim scholars denoted S is the logic corresponding to the following fragment of English:

Every p is a q: $\forall x(p(x) \rightarrow q(x)) \forall(p,q)$
Some p is a q: $\exists x(p(x) \wedge (x)) \exists(p,q)$
No p is a q: $\forall x(p(x) \rightarrow \sim q(x)) \forall(p,\sim q)$
Some p is not a: $\exists x((x) \wedge \sim q(x)) \exists(p,q)$

As this system is very well understood since ancient times and has been investigated recently in the shed of modern advances in formal logic, comparing its expressive power to FOL subsets used in computer science and investigating the complexity of algorithms based on it (cf. [Lukasiewicz]), emphasis in this paper shall be put on Qg'' which is the set of axioms of the logical theory being formalized and processed, i.e., *Fiqh*-axioms and rules[27]. Thus for the sake of simplification but without loss of generality, Qg = Qg'' in what follows.
Reaching a conclusion C through the (syntactical) rules of Qg will be depicted as the symbol ♦Qg C.
We then write: ♦Qg $(Fs \rightarrow Fv)$ to indicate that the sentence $(Fs \rightarrow Fv)$ has been deduced from the sentences Sen. 1, Sen. 2, Sen. 3.

---

[26] The notation A x B x C ... etc. is used for Cartesian products

[27] Note that the set of rules Qg'' in the previous example = {Sen1, Sen2, Sen3}.

## 9. SEMANTICS OF LOGIC SENTENCES

On the basis of the foregoing we conclude that sentences are initially of two types:

The first type contains variables of the type Qv, called *general sentences*, such as Sen. 1 and Sen. 2.

The second type has no variables. It only contains predicates and logical conjunctions, such as Sen. 3. We will call this type *detailed sentences*[28].

There is a marked difference between these two types and it clearly appears when we want to express the semantics, i.e., the values "true" or "false".

Let CS represent the set of true sentences in the formal system. It is clear that it will not be difficult for us to generate the meanings of the detailed sentences on the basis of CS, even if they are logically conjoint, because we know the meanings of symbols through QLn and also the meanings of the logic conjugations between these symbols. However, when we try to establish the meaning of the following formula, we are faced with some important problems:

Sen. 2: $((Fs \wedge Pgv \rightarrow X) \vee (Fs \wedge \textit{inverse}(Pgv) \rightarrow X)) \leftrightarrow (Fs \rightarrow X)$

We will deepen two of these problems:

The first is that this sentence has a general purpose, which means that its origin is:

---

[28] In mathematical logic those sentences are called "instantiated". We prefer here the term "detailed" to provide more clearer context to *Fiqh*-questions

For every X from QLn: $((Fs \wedge Pgv \rightarrow X) \vee (Fs \wedge \textit{inverse}(Pgv) \rightarrow X)) \leftrightarrow (Fs \rightarrow X)$

As such, it allows formulae such as:

Sen. 4: $((Fs \wedge Pgv \rightarrow Fs) \vee (Fs \wedge \textit{inverse}(Pgv) \rightarrow Fs)) \leftrightarrow (Fs \rightarrow Fs)$

In natural language: "I'tikaf is an essential condition for I'tikaf whether with or without a pledge, and all this is a reason because I'tikaf is an essential condition for itself".

This statement yields nonsense.

Should X therefore be limited to certain cases of the set QLn, or should it be left to include it all? If we were to limit X, what would be the procedure to be followed for this purpose?

The second problem lies in the fact that if we were to consider the sentences that yield new statements (or if we were to allow X to derive its values from CS instead of QLn) we would encounter logic formulae that involve circularity yielding even more nonsense, e.g.:

If the following predicate is true $M \equiv$ for every X from CS *inverse* (X)

M claims that all predicates from CS are false, but it is nevertheless one of these predicates.

When we replace X by the value M we come up with the *inverse* $(X) \equiv$ there is a true X from CS (X).

This contradicts the assumption that M is true meaning that the fact that M is true leads to being false, which is contradictory.

If we were to establish a pyramidal structure (hierarchy) so as to prevent general sentences from being uncontrollably mixed with other sentences and to prohibit in particular any definition that involves circularity, we may avoid such nonsense statements.[29]

Therefore, mathematical logic works with two sets:[30]

1. *SymG*   the set of the meanings of predicates
2. *VV*     the set of the values of variables

SymG includes all the predicates from the set QLn with the values "true" and "false". For example:

SymG ≡ (Fs = *true*, Pgv = *true*, Prv = *true*, Fv = *true*)

Let us express the case of a person who has performed I'tikaf, pledged, fasted and prayed all at the same time. If we define the set SymG in this way, the meanings of the detailed expressions such as (Fs ∧ Y) will easily be defined as we understand the meaning of the symbol ∧.[31]

However, if we want to know the meaning of:

For every X from QLn: ((Fs ∧ Pgv → X) v (Fs ∧ *inverse* (Pgv) → X)) ↔ (Fs → X)

it is necessary to restrict the variable X with one value (at a time).

This is done by using the set VV.

Let us say VV ≡ (X=Fv, X= Prv). We thus get:

((Fs ∧ Pgv → Fv) v (Fs ∧ *inverse* (Pgv) → Fv)) ↔ (Fs → Fv); and
((Fs ∧ Pgv → Prv) v (Fs ∧ *inverse* (Pgv) → Prv)) ↔ (Fs → Prv)

This shows that individual sentences yield different meanings according to the variation of the pair (SymG, VV) applicable to the sentence.

From this perspective, sentences are classified into true sentences for any possible pair (SymG, VV)[32] and are called *Valid Truth*. Evidently, there are sentences that do not meet this condition. In other words, there will be at least one pair of (SymG, VV) that yields the value "false".

We assign the abbreviation {(SymG, VV) ♥ Sen} if we wish to say that the pair (SymG, VV) assigns sentence Sen the value "true" and say "The pair (SymG, VV) *validates* Sen".

## 10. FORMAL INCOMPLETENESS

We now look into the relationship between formal (syntactical) deduction ♦ and model-based deduction ♥[33].

For any formal system which yields for every Qg=(Sen1.... SenN) and for every M from CS:

---

[29] This pyramidal structure shall have steps that are numbered from 0 to infinity, with the step 0 representing expressions that include characteristics of predicates, while the logical systems of step 1 include such expressions, and step 2 includes characteristics of these expressions, and step 3 characteristics of these characteristics, etc.

[30] Standard textbooks on mathematical logic use the slightly more complex concepts of "assignment" and "interpretation" to describe *correctness* and *completeness* in a formally correct way. The description used here is informal and intends to clarify those two concepts intuitively without going into the unnecessary details of a mathematical formalism.

[31] The meaning here is "true" because both Fs and Y are true predicates according to the said SymG.

[32] Such as the sentence Fs → Fs

[33] The relation ♥ is sometimes called "semantic entailment relation"

♦ Qg M ⇒ there is a pair (SymG, VV): (SymG, VV) ♥ M

Such a formal system is said to be *correct (=consistent)*.

And if it yields for every M from CS every pair (SymG, VV):

(SymG, VV) ♥ M ⇒ there is a set Qg= (Sen1... SenN): ♦ Qg M

Such a formal system is said to be *complete*.

Therefore, the equality of the two relations ♦ and ♥ would yield a complete *and* correct formal system.
Fundamental mathematical results show that even Arithmetic cannot be described using a complete formal system[34].
Some conditions rendering formal systems inconsistent/incomplete and relevant for the current work shall be summarized as:

1. Circular definitions of logic sentences (which we have already exposed). They lead to *inconsistency* because the correctness of a predicate is necessarily linked to its incorrectness.

2. The use of semantic sets with non-denumerable cardinalities, such as those required by general mathematical logic functions.[35]

The principle of incompleteness of FOL is considered one of the most important principles of contemporary computer science as it directly impacts the methods of general programming. Nevertheless, complete (and decidable)[36] formal systems with less expressive power than FOL are known to exist and can be used in the context of *Fiqh*:

## 11. COMPLETENESS AND DECIDABILITY OF FIQH-SYSTEMS

Some famous formal systems which are weaker than FOL, but both *complete* and *decidable* are among others: Propositional calculus, the set of FOL validities in the signature with only equality, the set of FOL validities in a signature with equality and one unary function, and the set of first-order Boolean algebras. It is also known that every complete recursively enumerable first-order theory is decidable[37].

---

[34] This famous result is attributed to *Kurt Goedel*'s incompleteness theorem (cf. [Goedel]).
[35] The notion "denumerable" sets (cf. [Cantor]) is related to the important notion of "constructability" of an algorithmic approach, i.e., all cases of problems where the underlying domain sets are not denumerable yield *undecidability* results (cf. fn 36)

[36] The term *decidable* refers to the *decision problem* (*Entscheidungsproblem*) which imposes the question of the existence of an effective method for determining membership in a set of formulas, or, more precisely, an algorithm that can and will return a Boolean true or false value (instead of looping indefinitely). Logical systems such as *propositional logic* are decidable if membership in their set of logically valid formulas (or theorems) can be effectively determined. A theory (set of sentences closed under logical consequence) in a fixed logical system is decidable if there is an effective method for determining whether arbitrary formulas are included in the theory. *Decidability* should not be confused with *completeness*. For example, the theory of algebraically closed fields is decidable but incomplete, whereas the set of all true first-order statements about nonnegative integers in the language with + and × is complete but *undecidable*.
[37] This does not contradict the famous incompleteness result, since it only says that one can start with any consistent first order theory T and construct a "logic completion" of T by

As seen earlier, syllogisms were used by ancient Muslim scholars to express and use rules of *Fiqh* in real world settings (c.f. e.g.: [Al-Akhdari])[38]. The completeness of various formulations of syllogistic logic has been demonstrated in [Lukasiewicz]. Although syllogistics provide with some quantification properties, there is a lack of predicates which have -arity many more than one as well as the possibility to consider functions. Thus they can be seen as subsystems of monadic first-order logic (MFO), which is also less expressive than full FOL.

It is worth noting that a thorough study of the formal systems for *Fiqh* needed, taking into account the unique formal properties of this domain, may result in the choice of any of the above restrictions of FOL. Such a study is due. Referring back to *Fiqh*, we can now discuss how to represent the previously suggested characteristic of *Consistent* and *Complete Legalization* by using formal systems.

For the set of *Fiqh*-questions we will use the symbol SF, for the set of rulings QR, and for the set of reasons QE respectively.

Thus SF ≡ (M1, M2, M3...Mn), QR ≡ (R1, R2, R3...R,), QE ≡ (E1, E2, E3...E,)[39]

*Complete Legislation* means, as we have already noted, that every religious question has a ruling (verdict). This means:

For every M of SF: there is R of QR: (M → R)

*Consistent Legislation* means that every religious ruling has a reason. This translates into:

For every M of SF and for every R of QR: ((M → R) → there is E of QE: (E → R))

We may choose

QLn ≡ QE ∪ SF ∪ QR and Qc ≡ (→, ∧, inverse) and Qv ≡ (X1, X2... Xn) and Gmr and Qg as discussed in section 8. In other words, QLn incorporates the totality of the predicates of all three sets.

**Proposition 3:** If a formal system is chosen in such a way that (QLn, Qc, Qv, Gmr, Qg) is both logically complete *and* decidable, then the question of *Complete* and *Consistent Legislation* (*Fiqh*) is *decidable* (i.e., it is decidable whether Lemma 1 holds or not).

Proof: *Complete and decidable logic system* is equal to calculate the meanings of all possible logic expressions. As such, it is necessarily possible to also know the meaning of the expressions *Consistent and Complete Legislation*.

---

adding logical consequences of the formulas in T in a constructive way. *Incompleteness* states that this process cannot be done for sufficiently expressive T.

[38] The most apparent reason for choosing syllogisms to express *Fiqh*-rules by ancient scholars is represented by the fact that those logic systems were the only ones known at that time.

[39] El represents the predicate "The reason l is a valid religious reason". The same applies to Rl and Ml.

## 12. COMPLEXITY OF DEDUCTION IN FIQH-SYSTEMS

An important aspect with respect to any formal logic system, especially if the intention is to implement it on modern computer systems, is the complexity of its deduction procedures. As is the case with many interesting sub-sets of FOL which are complete and decidable: The best known deduction methods require *exponential* number of steps in the worst case. A prominent example of such systems is *Propositional Logic*.

Although the decision concerning the best formal system to use for *Fiqh* must be subject to a thorough investigation eventually yielding a tailored formal system specifically designed for *Fiqh,* we will consistently assume *Propositional Logic*. Accordingly, the following two trivial propositions related to general and detailed sentences of Section 9 hold:

**Proposition 4:** The deduction of detailed sentences (by substitution into general ones[40]) is efficient, i.e., in the worst-case *polynomial* with regard to length of sentences.

Proof: Straightforward substitution of values into a logic formulae using simple procedures of linear complexity in the length of formulas.

**Proposition 5:** Deduction of general sentences from a set of general and detailed ones is *NP-Complete*[41]

Proof: It is well known that general deduction - like the one needed to infer general formulas from a set of *Fiqh*-axioms - in *Propositional Logic* is an instance of a *SAT problem*[42].

Proposition 4 and 5 have a direct practical consequence for the efforts to realize *Fiqh*-deductions on contemporary computer systems: While the amount of computation needed for answering concrete queries/questions is relatively small if the *Fiqh*-sub-system used is complete, i.e., if there are *Fiqh*-rules to be applied to each query (queries = detailed sentences), deducing *Fiqh*-rules from others may be computationally challenging.

The algorithm described below accounts for both cases, i.e., while describing a procedure of automatically solving user-defined *Fiqh*-queries, it also describes a *heuristic* way of finding "generic *Fiqh*-rules" starting from a set of *possible queries* in the different relevant chapters.

## 13. ALGORITHMIC VS LOGIC PROGRAMMING

The computer is one of the most important applications for formal mathematical systems. Perhaps one of its most evident advantages is its factual presentation of logic, i.e., many abstract theories are materialized by computer systems yielding visible results.

Yet not everything can be described logically. There are meanings that can only be referred to by terminology systems, such as everything that is described by a series of values that may be infinite. Such series may not contain any logical structure and it may have

---

[40] Which are basically *Fiqh*-rules.

[41] cf. *NP-complete*. (2014, February 20). In Wikipedia, The Free Encyclopedia. Retrieved 20:58, February 28, 2014, from http://en.wikipedia.org/w/index.php?title=NP-complete&oldid=596284955

[42] cf. *Boolean satisfiability problem*. (2014, April 30). In Wikipedia, The Free Encyclopedia. Retrieved 21:01, May 2, 2014, from http://en.wikipedia.org/w/index.php?title=Boolean_satisfiability_problem&oldid=606502483

values which have no apparent pattern, as is the case with the arbitrary numbers (12, 155, 76.4).[43]

In order to accommodate non-logical structures, a method is needed which relies on representing the problems to be solved in terms of successive commands in form of series that start and finish in a controlled way (algorithmic programming). Here, logic only determines the number and order of the solutions. This method can represent any series, whether logical or illogical, e.g., the *Algorithm* for Islamic heritage law in *Kitab Al-Jabr wa-l-Muqabala*.

The most important features of algorithmic programming languages compared to their logical counterparts are:

1. Sentences do not have a direct logical structure[44]. In other words, the structure does not include predicates linked by standard attributes. It relies on the principle of a chain of commands (solution steps).
2. Semantics are not simply logic predicates being true or false. They are internal states of a machine according to *operational* semantics or applications of mathematical functions according to *denotational* semantics. They follow one another according to the order of commands in the series.
3. There is no mathematical or logical possibility to check the correctness of the results of general algorithms. Indeed it is not possible to predict whether the algorithms will stop or not after the start of the solution operation.

By contrast, logic-programming languages rely on a simple logic representation of problems and have the advantage of being able to predict the results of application when the logic system represented by those programs is both, *complete and decidable*.

Those facts are mentioned here to underline the importance of the algorithmic method for entering and resolving practical questions of *Fiqh*. Although *Fiqh* meanings follow incrementally in a logical way that admits formal definition, people's actions, represented by possible religious questions, are not necessarily subject to immediate logical implementation. Hence an algorithmic method is necessary in addition to the logical one to represent such cases.

## 14. ALGORITHMIC IMPLEMENTATION

The need for looking at *Fiqh* in a formal logic framework is complementary to the urgent need of re-writing the *Fiqh* reference works in a style that can easily be understood by contemporary readers. Moreover, the re-writing should address the need to provide solutions for practical problems faced in an increasingly complex and differentiated world. Among others, this attitude is supported by *Al-Zuhayli* (1350AH- / 1923AD -):

> There is nothing to preclude re-writing Islamic *Fiqh* in the form of articles that make it easier for a judge, lawyer or ordinary person to refer to its rulings by article and paragraph. This is indeed one of the urgent needs of our time. Old books of *Fiqh* are not easy to refer to for a ruling on a particular question, except by a specialist. Indeed the whole body of *Fiqh* needs to be re-organized and re-arranged into

---

[43] e.g., most statistical and mathematic denotations as well as the digit distribution of all members of the set of real numbers (*per definitionem*)

[44] although they can be translated to non-intuitive logical sentences when needed

chapters, in accordance with the systems that are familiar to today's students and scholars. It is well known that *Fiqh* presents its material in detailed questions. It does not follow the approach of explaining its general theory before studying its contents. This presents real difficulties in identifying the religious rulings. (cf. [Al-Zuhayli])

While *Al-Zuhayli* stresses the need for re-arranging *Fiqh*-chapters he lags proposing or pointing to any specific method. The re-organization of legacy systems, however, has become a full-fledged discipline which is directly related to computer science incorporating new approaches that were unknown and impossible in earlier times. Legacy methods of chapter arrangement relied on a library system that arranged manuscripts on bookshelves according to their main or subsidiary headings. Students could only do research with a single indexing feature while enhanced methods classify *Fiqh*-information with combinatorial and semantic variety providing with more efficient research experience.

The here suggested classification is an inherent result of the algorithmic approach taking the user experience an important step further by re-arranging the books using all possible combinations of *Fiqh*-questions generated from sets of terminology trees valid in different chapters and/or sub-chapters allowing thus to address a question directly without assuming its answer beforehand.

## 15. PROGRAMMING METHOD

The method relies on generating all possible questions under any *Fiqh*-chapter or sub-chapter by means of an algorithmic machine. To translate the result of formal completeness and decidability into a practical application, the following steps were necessary:

(1) Define the formal structure of a *simple Fiqh*-question in the following way:

The algorithmic process of induction yields a religious question[45] composed of four essential elements:

1. subject
2. tool
3. reason
4. method

In case for example of a (non-trivial) human subject, initial kind-specific qualities for purification may consist of:

1. anatomy
2. conviction (religious affiliation)
3. properties
4. action
etc…

These qualities may interact so as to produce numerous possibilities and complex interactivity. Moreover, the main qualities may contain secondary qualities which may also interact within or outside the main qualities. Hence, a *compound Fiqh*-question[46] is

---

[45] Such as: a man performed *wudu* using water that has already been used to remove simple impurity.

[46] Such as: a man performed *wudu*, then urinated, then washed his private parts, then performed ablution.

(1) a conjugation that gives a ruling based on a sequence of *Fiqh*-actions/events generating previous ones in the light of the conditions of actions being linked to what is essential, recommended, etc.

(2) Construct a tree of defined terms for each question element which is devised to contain all aspects of the addressable *Fiqh*-questions (Figure 3 with a tree sample for all four elements of a classic *Shafi'i*-school *taymammum* sub-chapter).

(3) This stage is followed by what may be called *generation of practical questions*. As illustrated in Figure 3, a potentially large number of possible questions is generated by a straightforward combinatory algorithm. As trees get more detailed in describing the different aspects of a *Fiqh*-decision in the particular chapter, the combinatory becomes enormous[47].

(4) To reduce the combinatory explosion seen in step 3 practical rules are derived by excluding all possibilities providing algorithmically of null-results, such as the performance of *wudu* by a baby or the performance of *taymammum* without reference to a valid reason. This process yields *negative* practical rules while the complements (inverses) are generated on-the-fly as *positive* practical rules. The process of deducing negative and positive rules is an efficient practical (heuristic) alternative to classical deduction of general formulas which may be as seen in proposition 5 computationally intensive. This process is explained in detail in the next section.

(5) Based on this state of formal completeness the next and last stage consists of implementing a *complete* algorithm which contains the formalized, logic *Fiqh*-theory. This algorithm comes with two basic *flavors*: one which answers simple questions and one which answers compound questions as seen above. An example of the purification chapter, i.e., *Taharah*, will illustrate this step:

## 16. PRACTICAL INSTANCES OF FIQH-RULES (NEGATIVE AND POSITIVE)

The formal concept of negative and positive *Fiqh*-rules was introduced to reduce the apparent combinatory explosion resulting from step 3 above. Figure 4 illustrates the revision procedure needed to generate those rules. Truth table entries are inspected for the possibility of (manual) abstraction of rule patterns. The result is two sets of negative- and positive rules substituting the entire truth table[48].

Having exemplarily referred to *Al-Suyuti* (849-911AH / 1445-1505AD) (cf. [Al-Suyuti]), 12 applicable *Fiqh*-rules were divided during our practical work into negative and positive ones. Some of them are not applicable to one chapter only but to many. For some of the generated abstractions we could not find known primary rules in the *Shaffi'i*-

---

[47] The constructed trees for *taymammum* (as per *Shafi'i*) yield an amazing potential combinatory of 15G questions (i.e., 15 Billion questions) while the same number was 2000G questions for the whole *Tahara* book.

[48] This is at least the objective of this abstraction exercise. It may happen that parts of the truth table (i.e., single questions) do not fall under any category (whether +ve or −ve) of abstraction.

school indicating a gap in the ancient scholars descriptions of their used intrinsic logic[49].

Processing the rules related to purification ultimately generated a number of secondary rules:

Negative rules:

1. *Tayammum* becomes invalid when water becomes available. This rule is derived from a primary rule that says: "What is permissible as a result of a certain reason becomes invalid when that reason no longer applies."

2. Nothing other than water removes an impurity that is beyond a person's private parts. This is derived from a primary rule that says: "When the two rulings of permissible and forbidden apply to the same thing, the forbidden takes precedence." It is also possible to say that the use of a solid object, such as a stone, is permissible for the removal of the impurity from the private parts as a matter of necessity. This relies on the rules: "Necessities permit what is forbidden," and "Necessities are measured according to need."

3. It is not permissible to use something of value or something eatable to remove the impurity from one's private parts. This is derived from a primary rule that says: "Concessions may not be exercised by what is forbidden".

4. It is not permissible to use water that has been used (no primary rule).

5. *Tayammum* to perform an obligatory prayer may not be done before that prayer is due. This rule is a part of a rule mentioned by *Al-Suyuti*: "*Tayammum* to perform an obligatory prayer may not be done before that prayer is due and it cannot validate recommended prayers". This rule mentioned by *Al-Suyuti* is derived from a primary rule that says: "If a special case is no longer valid, does the general case apply?" (cf. [Al-Suyuti])

6. Sparkling water to remove the impurity caused by a girl's urine is not valid. (No primary rule).

7. In the case of *taymammum* the intention statement "I intend to perform *taymammum*" is not valid. This rule is an exception from the rule that says: "Whatever is derived from an intention statement that applies to a section of Islamic law should be expressly clear."

8. When a sufferer of incontinence is cured, his prior ablution is invalid. (No primary rule).

Positive rules

1. When the lesser impurity (i.e., *hadath*) and ceremonial impurity (i.e., *janabah*) apply together, the grand ablution (i.e., *ghusl*) is enough to remove both.[50]

2. When the ceremonial impurity (i.e., *janabah*) and menstruation apply together, one grand ablution (i.e., *ghusl*) is enough to remove both.

---

[49] It is here worth noting that such gaps – if confirmed – indicate a breach of proposition 1 and thus the *Fiqh*-consistency assertion for the *Shafi'i*-school (related to this specific chapter) and would require from *Shafi'i* scholars a revision of *logic* reasons and/or intentions behind those gaps. Negative- and positive rules are thus a very important tool to investigate completeness of *Fiqh*-theories.

[50] The lesser impurity occurs as a result of a discharge through one's private parts and it is removed by *wudu* while the ceremonial impurity occurs as a result of an ejaculation or sexual intercourse. It requires a full bath.

These two rules are derived from a primary rule that says: "When two matters of the same type apply and their purpose is the same, they become mostly concurrent."

3. The discharge of semen does not make *wudu* necessary. This is derived from: "What necessitates a greater action in particular, does not require the lesser one in general".

4. It is sufficient to use a dry object to remove the impurities of *madhi* and *wadi*. This is derived from the rule: "Should what is rare be treated separately or attached to what is of its kind?"

Transferring this theoretical information into practical *Fiqh* by computing all possibilities that can be derived from the set of terms yields the complete set of secondary questions, i.e., an intermediate step towards *Fiqh*-completeness as discussed above.

In contrast, a traditional *Fiqh*-scholar without computer-assisted aid cannot consider the totality of secondary questions, hence, remains in an intuitive, artistic anticipation of the most balanced ruling.[51]

## 17. REALIZED ALGORITHMS

Algorithms were realized in conceptual prototype terms for two different *Tahara* sub-chapters: *Wudu* and *taymammum*. For *taymammum*, both flavors of the method described herein were realized. The one replying to simple questions and the one replying to compound questions. For *wudu* only the compound question part was realized. The User Interface (UI) was implemented for Arabic as its translation to any other language required consensus on what terminologies to use for different *Fiqh*-expressions (nomenclature). In what follows only the simple question answering procedure for *taymammum* and the compound question answering procedure for *wudu* shall be described.

The conceptual schematic of the simple questions answering procedure is depicted in Figure 5. The method generated a complete truth table which included all possible *Fiqh*-questions in the sub-chapter of *taymammum*. The truth table was then reduced (manually) to two sets of rules which formed the backbone of the logical engine. A restricted logic compiler understanding the format of *Fiqh*-rules (which is that of propositional calculus formulas) was written (in Visual basic) to enable matching of User queries with rules and providing intelligent results (i.e., results with explanations) as illustrated in Figure 5.

Based on the principle of compound questions, the general *wudu*-algorithm shown in Figure 6 was also realized. As the main difference between the two flavors (simple and compound) is the fact that sequences of actions have to be taken into account in the latter case, the algorithm was realized in form of a finite state automaton[52]. Finite state automata have exactly the required expressive power to model sequences of *Fiqh*-related actions/events.

---

[51] This is not due to the difficulty of the task, but because the mere magnitude and complexity in terms of combinatory. Hence, continuous follow up is required to place such secondary questions under a general- or specific rule and treat them with suitable rulings.

As illustrated in Figure 6, this algorithm takes into account the following six obligatory actions to be performed in order to reach a valid state of *Tahara* (*Shafi'i)*:

- Correct intention assertion
- Washing the face
- Washing the arms up to the ankles
- Wiping parts of the head
- Washing the feet including the heels
- Doing all this in the above given sequence

As preserving the order of one's actions is required by the *Shafi'i*-school, the realized automaton is a deterministic one[53]. The UI provides action buttons for actions from 1-5. All these actions affect the state of purification where the algorithm changes the cases step-wise as the user enters another action. If a user misses to enter an action or if he/she changes the order of some action(s), the system is able to advise and correct him/her (pre-differentiation). The process of *wudu* can also be conjugated with its invalidation (*hadath*) algorithm generating rulings on whether the *wudu* remains valid or has been invalidated by some action/event sequences. This has been realized in the *taymammum*-version of the compound algorithm and is not described here.

---

[52] cf. *Finite-state machine.* (2014, May 3). In Wikipedia, The Free Encyclopedia. Retrieved 21:18, May 3, 2014, from http://en.wikipedia.org/w/index.php?title=Finite-state_machine&oldid=606933991

[53] A *Hanafi wudu*-Automaton is not necessarily deterministic since preserving the order of *wudu*-actions is not obligatory according to the *Hanafi*-school.

## C. RESULTS AND DISCUSSION

In 1928, the German mathematician *David Hilbert* (1862-1943) formulated a mathematical question in terms of the *Entscheidungsproblem*. Its essence asks for an algorithm to decide whether a given statement is provable within the set of axioms using the rules of logic (cf. [Church] and [Turing]). Although the narrow context of this scientific project was to try to solve important questions of fundamental mathematics, its broader context was much more important, namely: To formalize a generic tool for informed, precise and transparent decision-making.

Muslim philosophers and scholars didn't go a much different path in the dawn of Islamic civilization. As mentioned in the introductory remarks, fundamental properties of Islamic Law (*Shariah*) and its *Fiqh*-sub-systems were investigated with such a precision that inventing new, previously unknown methods and concepts to enable answers to *Fiqh*-questions became necessary. In this context, fundamental mathematics played a vital role. Not that much as a subject of investigation but rather as a means to study the newly created *Fiqh*-machinery. Muslim scholars knew that they needed to be sure that their emerging, previously unknown tool of *Fiqh* was precise enough to answer even the most sophisticated questions. Although their *Entscheidungsproblem* was not explicitly formulated (asking for an algorithm to decide whether a given statement is provable from the *Fiqh*-axioms using the *Fiqh*-rules), it was implicitly understood that they were working in the direction of a positive solution[54].

---

[54] Many ancient *Fiqh*-books contain claims of completeness of sets of generic *Fiqh*- or *usul*-rules. Some of the scholars used to reduce basic principles of their school to only a small amount

As such, it is not entirely new to express an *Entscheidungsproblem* for *Fiqh* and to determine criteria under which it can be positively solved (cf. Proposition 3). Pre-requisite for doing so was the formulation of the properties *consistent* and *complete* for *Fiqh*-Legislations (Propositions 1 & 2).

Since the attribute of *completeness* depends on how far *Fiqh*-rules and their reasons can fulfill the underlying formal principles, it is evident that formally *complete* legislations of *Fiqh* were never discussed before, neither by leading scholars of *usul* nor by ancient Muslim scholars of logic. No matter how comprehensive a scholar's ability to memorize, classify and induce, he cannot claim to have looked into all possible questions within a certain area and verified them all consistent with the rule applied to them. Today, we are in a much more comfortable position as we can leverage machines deploying logic. With the help of these machines we can consider looking into all these possibilities, at least the ones related to a particular set of terms.

*Complete Legislation* appears certain at first sight,[55] but a careful examination shows that there is no extensive study of how far the *usul*-rules stated in scholars' books give all rulings. Hence, attention should be paid to regulate this aspect *mechanically*, as manual regulation is *practically* impossible.

## 18. SUMMARY

The results of this work shall be summarized as follows:

One: *Usul-* and Logic:

1- Consistency and completeness properties of *Fiqh*-legislations are formalized in mathematical logic terms (Propositions 1 & 2 as well as Lemma 1)
2- An *Entscheidungsproblem* is formulated with the criteria for its positive solution clearly determined (Proposition 3)
3- Complexity of *Fiqh*-algorithms (detailed- and generic queries) is investigated (Propositions 4 & 5)

Two: *Fiqh*:

1. Reclassification of the chapter on *Tahara* according to the *Shafi'i*-school converting it into secondary cases pertaining to questions that have the four essential elements

2. Drawing logical trees linking the four essential elements to *Fiqh*-decisions in the chapter

3. Introduction of negative and positive rules (+ve & -ve) formalized in propositional calculus form which are essentially a way of compressing *Fiqh*-chapters

Three: Application:

1. Use of *wudu/taymammum* prototype algorithms for applications that answer user-defined *Fiqh*-questions with no or very little *Fiqh*-background

2. Assistance to scholars by identifying negative and positive rules in a chapter to investigate *consistency* and

---

of those rules claiming their sufficiency to answer all related questions and provide a rational argumentation for that answer.

[55] It is commonly assumed for the *Islamic Jurisprudence* to give a ruling on everything.

*completeness* of the *Fiqh*-theory of a given *Fiqh*-school and identify logical gaps in classical *Fiqh*-arguments

Future objectives may be summed up as follows:

1. Fully leverage the work of earlier scholars defining many ancient concepts more precisely especially those related to *consistency* and *completeness*
2. Investigate syllogistic and propositional systems for their adequacy to be used as a basis for modern, algorithmic *Fiqh*-machines (especially SAT problems occurring in both systems and their possible solutions)
3. Revisit the current definitions of *usul-* and *Fiqh*-rules yielding a more precise logical framework
4. Re-arranging ancient books in a system taking questions/queries into account *instead* of themes only
5. Ensure that questions are classified properly under the largest controlled number of specific- and general *Fiqh*-rules
6. Explore new yet inherent *Fiqh*-rules
7. Check the applicability of *Fiqh*-rules to present and future cases
8. Ascertain how a *Fiqh*-rule truly operates while defining it in a solid way.[56]
9. Attain *Fiqh-completeness* for chapters and/or sub-chapters
10. Identify and put together all *Fiqh*-reasons mentioned in ancient books
11. Confirm the importance of logic in the study of *Fiqh*-limitations
12. Operate the process of analogy (*qiyas*) fully mechanically
13. Enable continuous *ijtihad* processing on 24x7 basis
14. Make *Islamic Jurisprudence* (all schools) available to audiences around the world by translating *Fiqh*-algorithms to all major languages
15. Get most balanced religious answers to people's current and future questions
16. Train students of Islamic Studies to give sound and most balanced rulings using formal *Fiqh*-systems
17. Establish a rational basis for correspondence between Islamic and secular legislations

Together with broadly built-up internet infrastructure, a re-algorithmization of *Islamic Jurisprudence* with an appropriate cloud-presence not only includes the possibility for access by the people, but with:

a) *Transparency* and *accountability* on the axiomatic level, i.e., which axioms were considered, how and why?
b) *Completeness* and *consistency* as discussed in the preceding chapters yielding a total consideration of axioms and rules.
c) *Legal security* and *legal equality* with *rule of law*, i.e., any verdict must account for the axioms which are completely transparent to the whole world, for Muslims and non-Muslims alike.

---

[56] Is the rule merely collation of secondary cases, or does it operate like the general rules? Are they flexible to accommodate new cases?

## 19. FUNDING

This research was supported by *Tekcentric Corporation*, 155 P. Moffett Park Drive, Sunnyvale, 94089 California, USA; and *Mir Amir Ali*, Belmont, 94002 California, USA.

## 20. ACKNOWLEDGEMENT

A special appreciation is dedicated to *Adil Salahi*, London (UK), who translated important parts of this paper (Arabic into English) from a previous concluded technical whitepaper (2001) and to *Sami M. Angawi*, Jeddah (KSA), who provided with critical philosophical discussions.

## 22. FIGURES

Figure 1

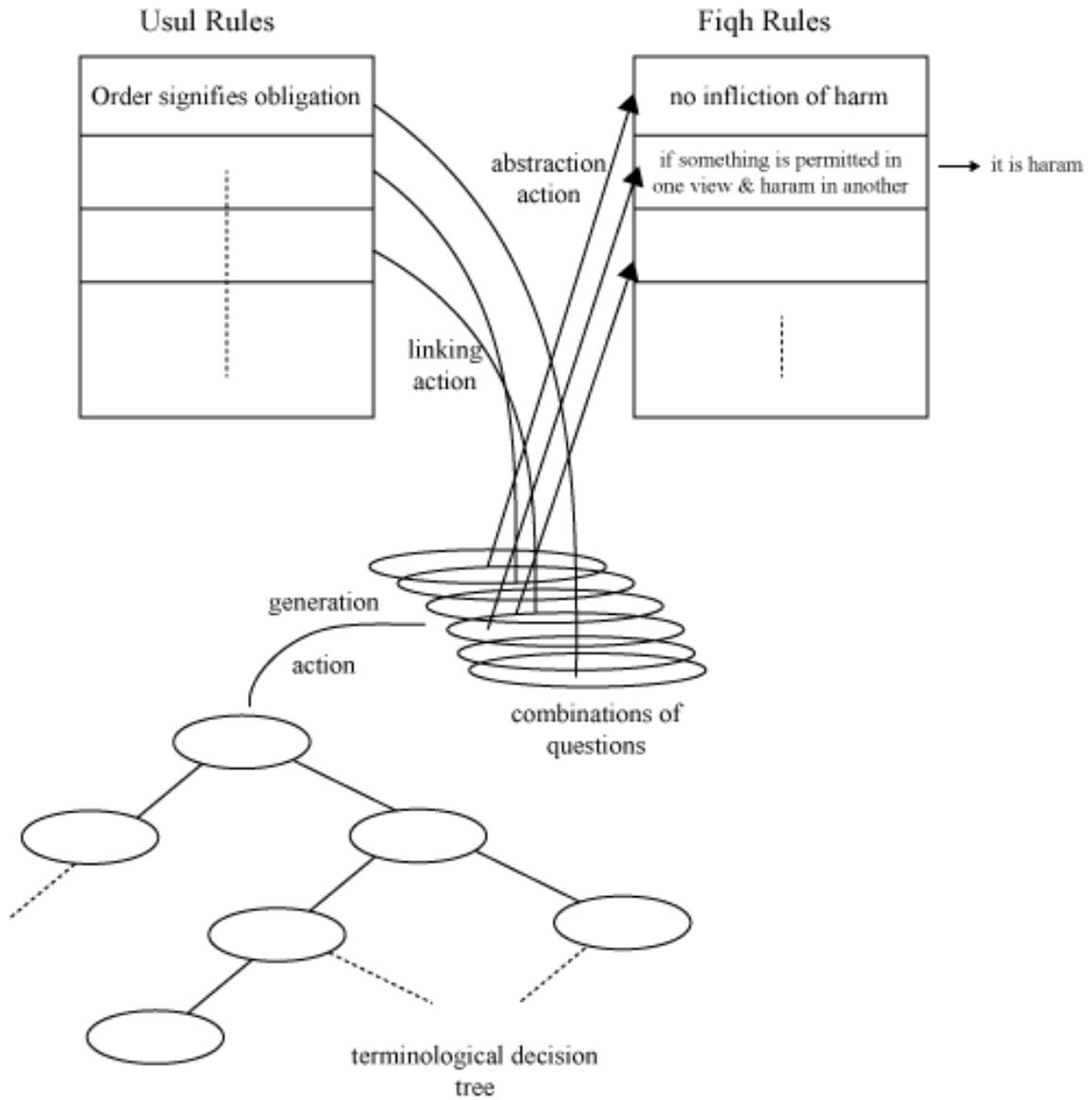

*Usul*-rules act as axioms for detailed rulings with the *Fiqh*-rules processing the logical implications to accomplish the axioms.

Figure 2

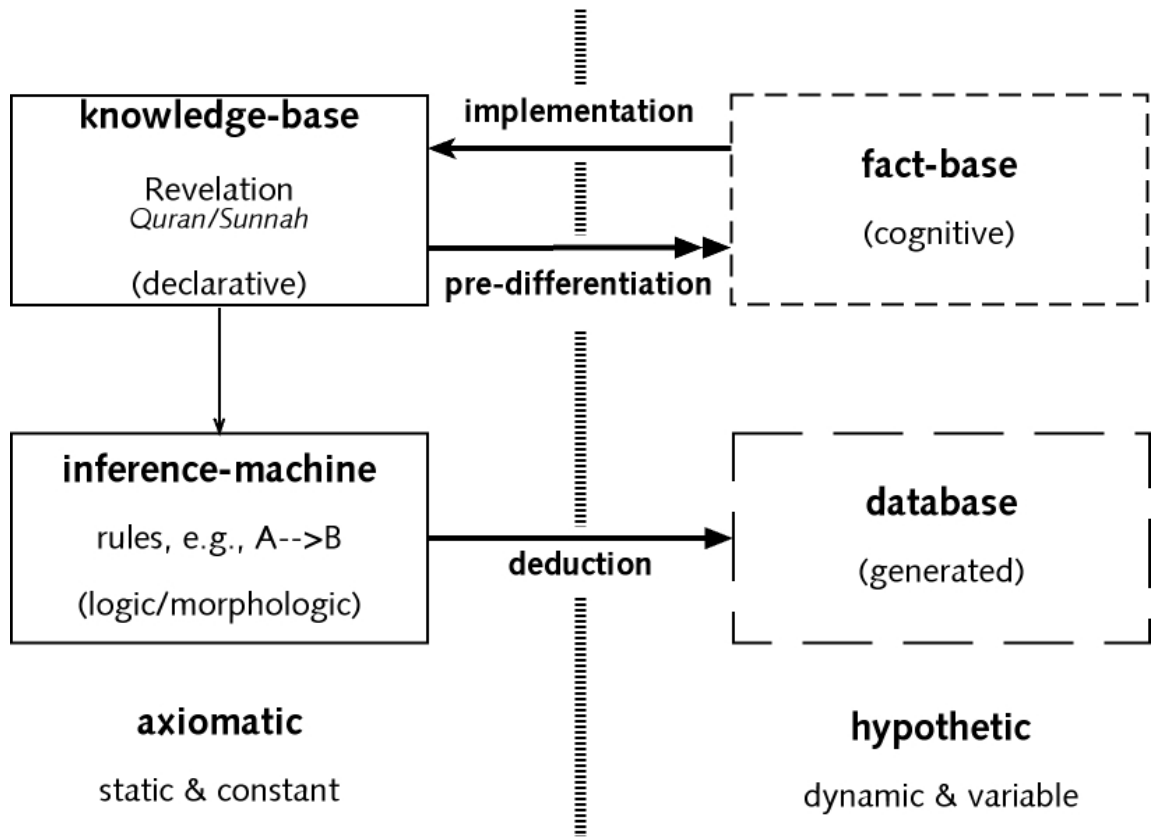

Together with primary sources (Quran and Sunnah), *usul-* and *Fiqh*-rules reside on the *axiomatic* side (left part). The fact-base and generated verdicts (database) reside on the hypothetic side (right part) yielding *balance* between *static & constant axioms* and *dynamic & variable hypothesis*. The process of pre-differentiation reduces complexity by heuristically increasing query-efficiency with a syntactically- and semantically closed terminology system defined by primary sources.

Figure 3

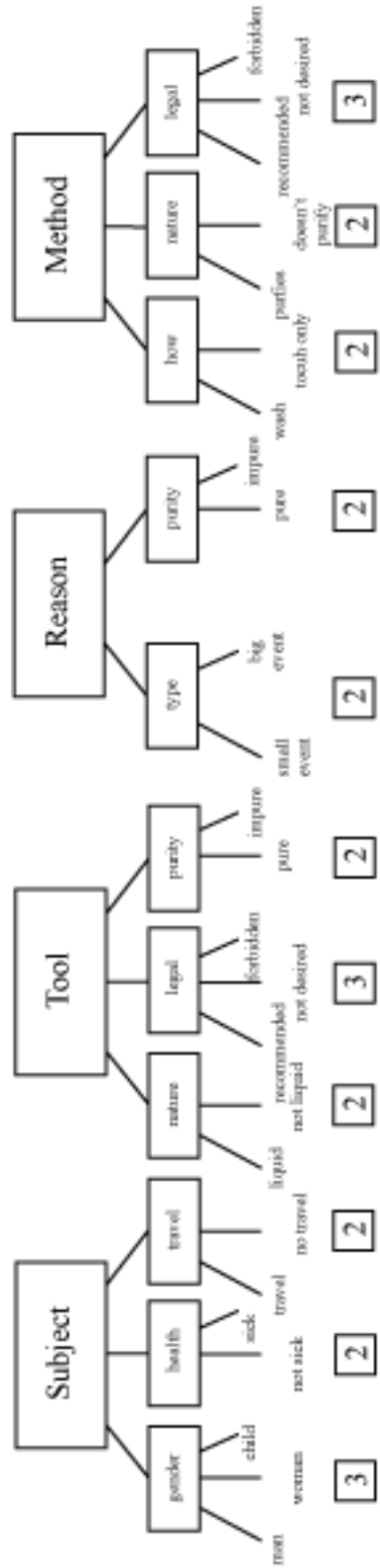

Terminological decision tree sample for question elements in *Taiammum* sub-chapters

Figure 4

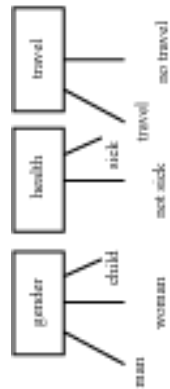

Figure 5

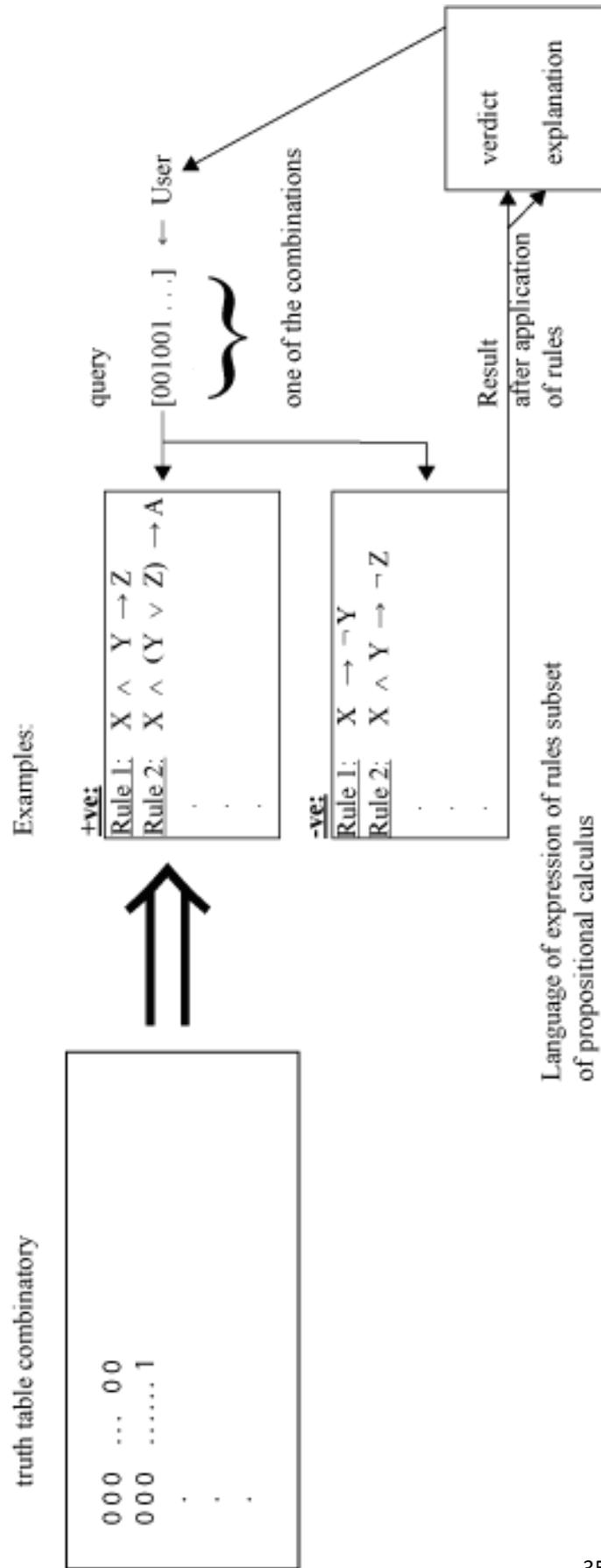

Figure 6

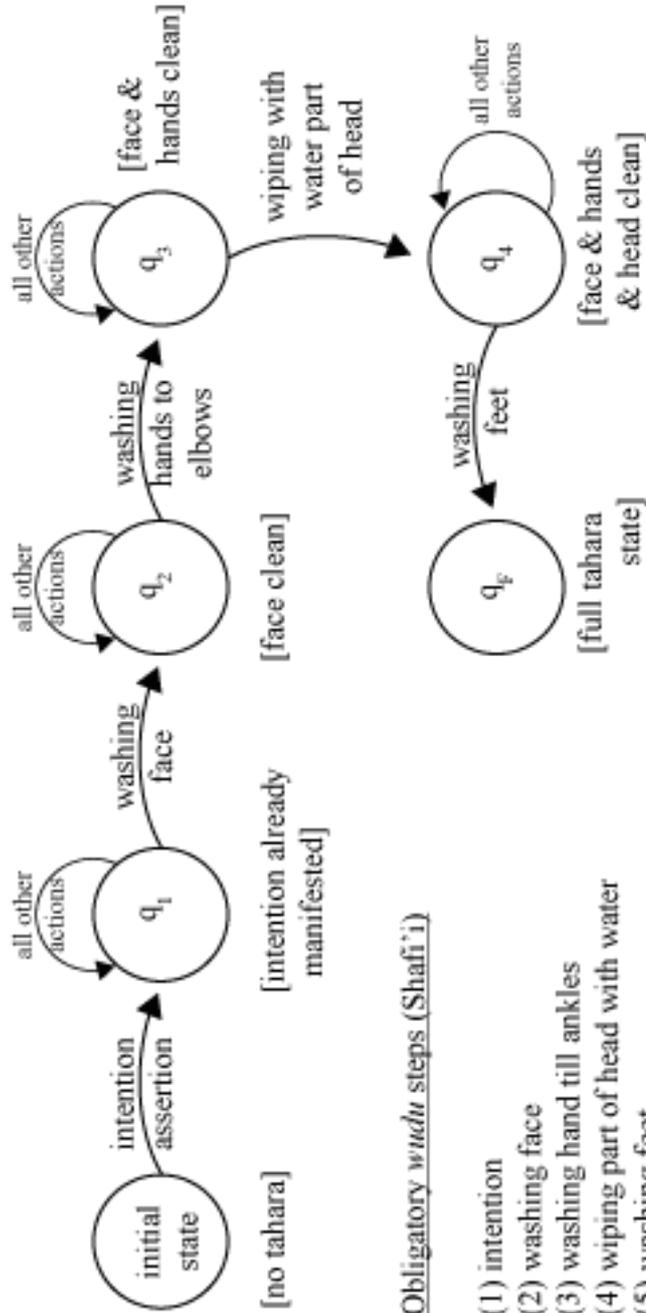

Answering compound questions (*Wudu*-Automation)

Obligatory *wudu* steps (Shafi'i)
(1) intention
(2) washing face
(3) washing hand till ankles
(4) wiping part of head with water
(5) washing feet
(6) doing all this in specified sequence